\documentclass[sigconf,nonacm]{acmart}
\usepackage{graphicx}
\newcommand{\datasetlink}{\url{https://huggingface.co/datasets/mws-physical-ai/dataforgevla-340}}
\newcommand{\skoltechname}{Intelligent Space Robotics Laboratory, Skolkovo Institute of Science and Technology}
\newcommand{\mwsname}{R\&D Center, MWS}
\newcommand{\correspondingemail}{sergei.kurchev@skoltech.ru}
\newcommand{\skoltech}{%
  \affiliation{%
    \institution{\skoltechname}
    \city{Moscow}
    \country{Russian Federation}
  }%
}
\newcommand{\mws}{%
  \affiliation{%
    \institution{\mwsname}
    \city{Moscow}
    \country{Russian Federation}
  }%
}

\makeatletter
\renewcommand{\@mkauthors}{%
  \global\setbox\mktitle@bx=\vbox{%
    \unvbox\mktitle@bx
    \hsize=\textwidth
    \centering
    {\large
      Sergei Kurchev\textsuperscript{1,2,*}\quad
      Iaroslav Kolomiets\textsuperscript{1,2}\quad
      Filipp Maksimov\textsuperscript{1,2}\quad
      Elizaveta Semenyakina\textsuperscript{1,2}\par
      Miguel Altamirano Cabrera\textsuperscript{1,2}\quad
      Artem Lykov\textsuperscript{2,1}\quad
      Dzmitry Tsetserukou\textsuperscript{1}\par
    }
    \smallskip
    {\normalsize
      \textsuperscript{1}\,\skoltechname\par
      \textsuperscript{2}\,\mwsname\par
      Moscow, Russian Federation\par
    }
    \smallskip
    {\small \textsuperscript{*}\,Correspondence:
      \href{mailto:\correspondingemail}{\correspondingemail}\par}
  }%
}
\makeatother

\begin{document}

\title[DataForgeVLA]{DataForgeVLA: Fully Synthetic Manipulation
  Demonstrations for Vision-Language-Action Training}

\author{Sergei Kurchev}
\email{\correspondingemail}
\author{Iaroslav Kolomiets}
\author{Filipp Maksimov}
\author{Elizaveta Semenyakina}
\author{Miguel Altamirano Cabrera}
\skoltech
\mws

\author{Artem Lykov}
\mws
\skoltech

\author{Dzmitry Tsetserukou}
\skoltech

\renewcommand{\shortauthors}{Kurchev et al.}

\begin{abstract}
We present a dataset of 340 fully synthetic manipulation demonstrations generated from real seed data for Vision-Language-Action (VLA) training. The corpus contains 32,340 frames with synchronized RGB and actions across eight object categories, represented as LeRobot v2.1 episodes with H.264 video, per-frame robot states and pose-valued action labels, language task annotations, summary statistics and quality decisions. Each synthetic demonstration is produced without additional teleoperation: a LoRA-fine-tuned Cosmos-3 model generates a manipulation video, and RoboHaMeR recovers the corresponding robot trajectory from RGB. Quality screening accepted 300 of 340 episodes for policy fine-tuning. As preliminary evidence of feasibility, a pretrained GR00T~N1.6 policy fine-tuned exclusively on the 300 accepted episodes reaches 16/20 (80\%) real-world pick-and-place success on the Oreo-package task. A policy fine-tuned on 105 real teleoperated episodes achieved 18/20 (90\%) under the same protocol. These results are based on one object and 20 trials per policy and are therefore reported as preliminary evidence rather than a general performance claim.
\end{abstract}

\ccsdesc[500]{Computing methodologies~Robotics}
\ccsdesc[300]{Computing methodologies~Machine learning}
\ccsdesc[100]{Computing methodologies~Computer vision}

\keywords{synthetic data generation, vision-language-action models,
  generative video models, robot manipulation, imitation learning,
  diffusion transfer, action extraction}

\begin{teaserfigure}
    \centering
    \includegraphics[width=0.93\linewidth]{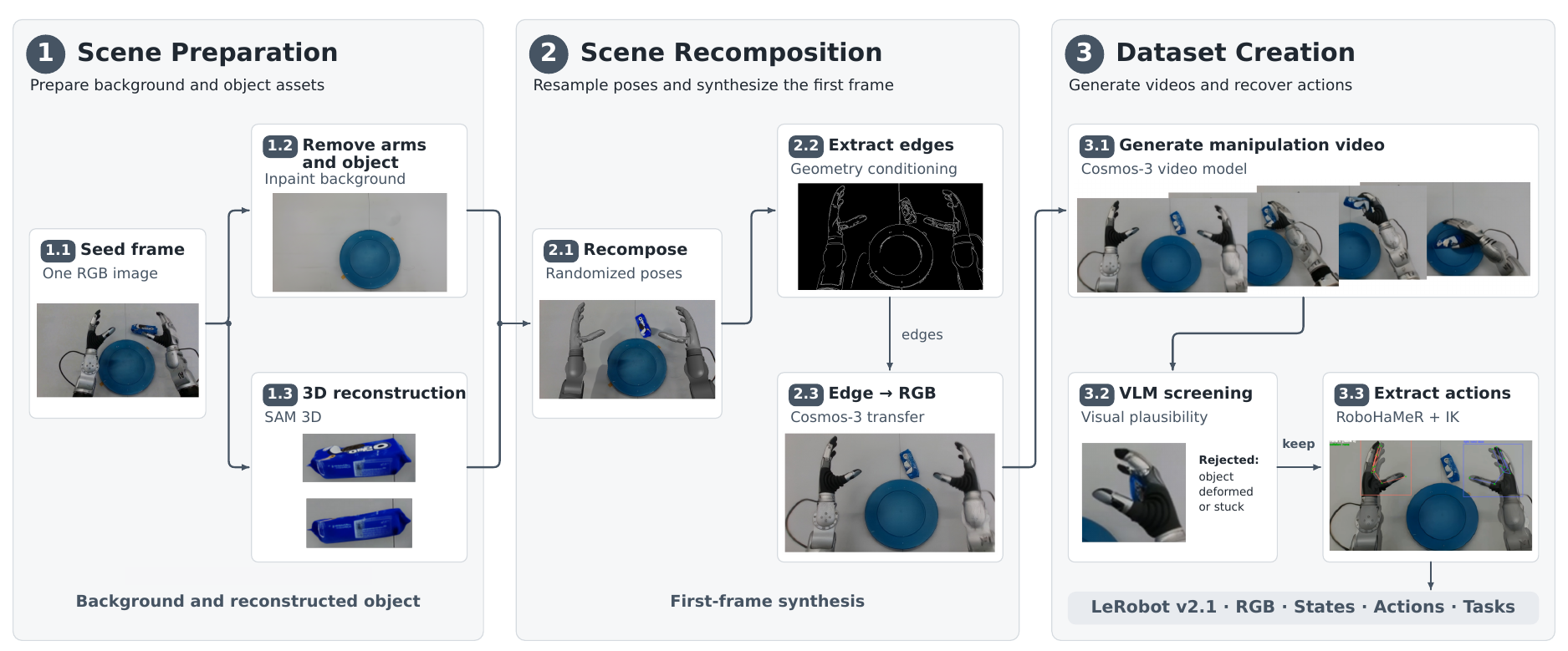}
    \caption{Overview of the DataForgeVLA generation pipeline. The quality gate
    selects clips for policy training. The public dataset also retains rejected
    clips and their action labels.}
    \Description{Diagram of the DataForgeVLA pipeline: a seed frame is
    emptied of the robot and object via segmentation and inpainting,
    recomposed with a resampled MuJoCo overlay and 3D object asset,
    converted to an edge map, passed through the Cosmos-3 transfer
    model to produce a photorealistic frame, animated by a Cosmos-3
    video model, and finally processed by RoboHaMeR to extract the
    action trajectory.}
    \label{fig:pipeline}
\end{teaserfigure}

\maketitle

\section{Introduction}
\label{sec:intro}

Vision-Language-Action (VLA) policies for robot manipulation improve with the diversity of the demonstrations they are trained on -- different objects, grasp poses, backgrounds, and task configurations. In practice, that diversity is bottlenecked by how the demonstrations
are collected: teleoperated data collection requires human operator time, and additional objects or scene configurations can require repeated
collection sessions.

A LoRA-fine-tuned Cosmos-3 video model~\cite{lora} can generate robot manipulation clips from a starting frame, while RoboHaMeR can recover robot trajectories from RGB without proprioceptive logs. Together, they turn generated clips into labeled demonstrations.

DataForgeVLA controls the starting scene by removing the hands and object from a real seed frame, then inserting a MuJoCo~\cite{mujoco} robot overlay and reconstructed object at sampled poses. A transfer model converts this low-fidelity composite to a photorealistic first frame before video generation and action extraction (Section~\ref{sec:method}).

The primary artifact is a documented, machine-readable corpus of synthetic RGB demonstrations paired with pose-valued action labels, together with a description of its generation and processing procedure. By generating additional manipulation demonstrations from a limited set of human-teleoperated episodes, our approach may reduce the need for repeated teleoperation across different robot and object configurations. This capability is relevant to HRI settings in which collecting demonstrations requires substantial human involvement and may entail operator fatigue and cognitive load.
Our contributions are:
\begin{itemize}
\item a LeRobot v2.1 dataset containing 340 episodes and 32,340 synchronized frames for eight object-conditioned pick tasks, including video, actions, language annotations, statistics, and per-sample quality decisions;
\item a description of a construction procedure combining controlled scene
  recomposition, generative video, and RGB-to-action extraction, reusing
  models trained on teleoperated demonstrations to create the
  generated demonstrations without additional teleoperation;
\item preliminary evidence -- a 16/20 (80\%) real-world success rate
  for a pretrained GR00T~N1.6 policy fine-tuned exclusively on the
  300 quality-screened episodes on one object -- that the extracted
  labels can support real-robot execution.
\end{itemize}

\section{Related Work}
\label{sec:related}

CACTI~\cite{cacti}, GenAug~\cite{genaug}, and ROSIE~\cite{rosie} alter the
appearance of objects, backgrounds, and distractors while retaining recorded
action trajectories. MimicGen~\cite{mimicgen} retargets demonstrations to new
object poses in simulation, which requires simulation assets and task
definitions and can introduce a visual and contact-dynamics sim-to-real gap.
Real2Render2Real~\cite{real2render2real} synthesizes robot trajectories from
tracked object motion, while DemoGen~\cite{demogen} adapts trajectories using
edited point-cloud observations.

Video world models can also generate the motion itself. DreamGen~\cite{dreamgen}
fine-tunes a video model on teleoperated data, generates video from an initial
image, and infers pseudo-actions with an inverse-dynamics or latent-action
model. Its humanoid evaluation reports
real-robot generalization after teleoperation of one pick-and-place task in
one environment. PhysicalAgent~\cite{physicalagent} uses generated video
during execution, proposing a new motion when the robot fails. Like DreamGen,
DataForgeVLA stores generated clips with recovered trajectories for policy
training, with explicit control over the initial robot pose and the object's
identity and pose through scene recomposition. AutodidactWAM~\cite{autodidactwam}
fine-tunes a world-action model's action branch with trajectories recovered
from its own videos. If a model trained this way can generate action labels
that support downstream policy training, it could supply both the clip and
its labels without a separate RoboHaMeR pass at generation time.

Human footage offers a source of motions that were physically demonstrated.
Phantom~\cite{phantom} uses hand-pose estimation and visual editing to turn
human videos into robot training data. AgenticFocus~\cite{agenticfocus} retargets
hand motion from first-person video to a humanoid while preserving the objects
in the resulting views. Retargeted human grasps could complement the motions
generated from recomposed scenes in DataForgeVLA.

Our pipeline uses Cosmos-3~\cite{cosmos3} for transfer and video generation,
GR00T~N1.6~\cite{groot} for policy training, and SAM~2~\cite{sam2} and
SAM~3D~\cite{sam3d} for hand segmentation and object reconstruction.
RoboHaMeR, a robot-specific adaptation of HaMeR~\cite{hamer}, recovers the
trajectories directly from RGB (Section~\ref{sec:robohamer}). Video artifacts
and extraction errors can affect the resulting labels.

\section{Dataset Construction}
\label{sec:method}

DataForgeVLA has two stages that are trained/used in sequence: first, an edge-to-RGB transfer model is trained once to make MuJoCo overlays photorealistic (Section~\ref{sec:method-transfer}); second, that transfer model is used inside a per-scene generation loop that produces new (video, action) training pairs (Section~\ref{sec:method-dataset}). Figure~\ref{fig:pipeline} summarizes both stages.

\subsection{Training an Edge-Consistent RGB Transfer Model}
\label{sec:method-transfer}

The goal of this stage is a model that, given an edge map derived from a MuJoCo-rendered robot overlay, synthesizes a photorealistic RGB frame with correctly posed, plausible robot hands -- so that later, synthetic overlays (Section~\ref{sec:method-dataset}) can be converted into training-quality images. Training pairs are built without manual annotation, by exploiting a calibrated camera and ground-truth robot poses recorded during ordinary data collection:

\begin{enumerate}
\item From a real frame, SAM~2 provides segmentation masks for the
  robot's hands; those masked regions are then removed by a separate
  inpainting model, leaving a background plate with no robot hands.
\item The ground-truth robot pose for that frame is rendered in
  MuJoCo and composited onto the inpainted plate. Because the camera
  is calibrated and the pose is ground truth, the rendered overlay
  aligns closely with where the real hands were.
\item The composite is converted to an edge map, discarding the
  overlay's (synthetic, low-fidelity) texture and keeping only its
  geometry.
\item The pair (edge map, original real RGB frame) is used to LoRA
  fine-tune the Cosmos-3 transfer model for 1,000 steps on the 105 teleoperated
  seed episodes, i.e., the model learns to map
  MuJoCo-overlay geometry back to the photorealistic appearance that
  originally produced it.
\end{enumerate}

Camera calibration and ground-truth joint configurations provide geometric alignment, making manual pose annotation unnecessary.

\subsection{Synthetic Scene Recomposition and Dataset Generation}
\label{sec:method-dataset}

Given the transfer model from Section~\ref{sec:method-transfer}, each
synthetic training example uses the first frame of one real
teleoperated episode as a tabletop scene source and an object asset as follows:

\begin{enumerate}
\item \textbf{Scene emptying.} SAM~2 provides segmentation masks for
  the robot's hands; those masked regions are then removed via a
  separate inpainting model, as in Section~\ref{sec:method-transfer}.
  SAM~3D reconstructs the Oreo asset from that frame and the seven
  other object assets from separate photographs. Further inpainting removes
  the Oreo package, leaving an empty tabletop background.

\item \textbf{Resampling.} A MuJoCo robot overlay is rendered at a
  newly sampled pose, and the extracted 3D object asset is composited
  at a newly sampled position and orientation, onto the empty plate.
  Because the object exists as a 3D asset and the robot pose comes
  from MuJoCo rather than a recording, this step is not limited to
  combinations that were ever physically observed together.
\item \textbf{Transfer.} The resampled composite is converted to an
  edge map and passed through the transfer model
  (Section~\ref{sec:method-transfer}), producing a photorealistic
  first frame for the new scene configuration.
\item \textbf{Video generation.} The synthesized first frame is fed to
  a Cosmos-3 video model (LoRA-fine-tuned for 2,500 steps on the 105 teleoperated
  seed episodes), conditioned on a short task
  instruction (e.g., ``pick up the object and place it in the center
  of the frame''). Cosmos-3 generates the clip at 30\,fps
  ($\approx$193 frames, $\approx$6.4\,s); the raw MP4 is stored at
  the native 30\,fps.
\item \textbf{Action extraction.} The generated clip is downsampled
  from 30\,fps to 15\,Hz (every other frame) before RoboHaMeR
  processes it; this yields 97 frames for the standard $\approx$6.4\,s clips (320 episodes) and
  65 frames for the 4-second variations (20 episodes). RoboHaMeR's predictions are retargeted
  through the robot kinematic model and stored as per-frame absolute
  robot states and actions.
\item \textbf{Quality annotation.} We screen each generated clip for overall consistency using the procedure in Section~\ref{sec:VLMfilter} and store the resulting decisions and scores in the accepted and rejected separate files.
\end{enumerate}

The output of this loop is a (video, action) pair that did not require additional teleoperation to produce. Repeating it with different sampled robot/object configurations allows the dataset to expand beyond the configurations physically observed in the seed data, rather than being limited to combinations that were directly collected. The generated corpus contains all 340 generated episodes, one per generation attempt. The VLM filter rejected 40 episodes and accepted 300 (88.2\% of attempts) for policy fine-tuning. The preliminary result below (Section~\ref{sec:results}) uses this accepted subset. The full corpus contains 200 clips of the Oreo package and 20 clips for each of apple, charger, drink, red ball, tissues, toothpaste, and toy.

One real tabletop frame and separate object photographs yielded 170
recomposed conditioning frames, each paired with two video seeds to produce
340 clips: 100 frames for Oreo and 10 per other object. The real source collection comprises
105 unique annotated teleoperated episodes used to train the transfer,
video, and action-extraction models.

\subsection{RoboHaMeR Label Generation}
\label{sec:robohamer}

RoboHaMeR adapts the pretrained HaMeR ViT-H encoder and transformer decoder~\cite{hamer} to the Unitree G1 with BrainCo hands. A hand detector supplies per-frame crops; left-hand crops are mirrored into a canonical hand frame. The network predicts six normalized actuator commands, a 6D representation of wrist rotation that is converted to $SO(3)$, and camera-to-palm depth.
Inverse kinematics retargets these to the G1 model; temporal tracking and a physical jump gate are applied before export. Trained on the same 105 seed episodes, the extractor achieves 3.57\,mm mean palm depth error, 0.734\,mm mean forward-kinematics keypoint error, and 0.0094 MAE on actuator commands on the held-out split. The keypoint error is measured over the 11 physical joints of the BrainCo hand in a root-relative frame, isolating articulation accuracy from global wrist positioning. While a 21-point MANO-compatible skeleton is extrapolated for 2D visual tracking, the 3D metric reflects pure joint-space prediction accuracy.

\subsection{VLM Quality Filter}
\label{sec:VLMfilter}

For each clip, we process up to 256 frames. The Cosmos3-Nano judge first evaluates the full clip, then evaluates four shorter, overlapping temporal windows. Each segment contains approximately 40\% of the sampled frames. We disable generation sampling and instruct the judge to return only \(1\) (accept) or \(0\) (reject).

The full-clip judgment checks whether the requested manipulation is completed with the correct object and target, plausible hand to object contact, and coherent motion. Segment judgments check object correspondence, shape consistency, and physical plausibility. These local checks help catch brief failures that may be overlooked when judging the clip as a whole, such as an object fragmenting, disappearing, or jumping during the action.

Let \(s_{\mathrm{clip}}\) denote the full-clip judgment and \(s_1,\ldots,s_4\) the segment judgments, each in \(\{0,1\}\). An episode is accepted only if all five judgments equal 1,
\[
\min(s_{\mathrm{clip}}, s_1, s_2, s_3, s_4) = 1.
\]
The decisions and model responses with scores are retained in separate accepted and rejected JSONL manifests.

\section{Dataset Description and Usage}
\label{sec:dataset}

The public DataForgeVLA archive is available at \datasetlink. It contains both the
46D kinematic corpus and the 13D right-side training export for all 340 generated
episodes, together with metadata, an MIT license, and quality-decision manifests
for 300 accepted and 40 rejected episodes. The archive also includes the 13D
converter, the training configuration, and the accepted episode IDs.

\begin{table}[t]
  \caption{Full 46D corpus and archive metadata.}
  \label{tab:dataset}
  \centering
  \small
  \begin{tabular}{@{}lr@{}}
    \toprule
    Item & Value \\
    \midrule
    Episodes / synchronized frames & 340 / 32,340 \\
    Object categories / task strings & 8 / 8 \\
    Video files & 340 MP4 (H.264, 640$\times$392, 15 Hz) \\
    Trajectory files & 340 Parquet \\
    Episode length & 97 or 65 frames \\
Metadata / quality files & 7 / 3  \\
    Official split & train: episodes 0--339 \\
Archive license & MIT \\
    \bottomrule
  \end{tabular}
\end{table}

The 46D Parquet records contain \path{observation.state}, \path{action}, timestamps, frame and episode indices, and task indices mapped to instructions in the metadata. These trajectories use a 46-channel bilateral kinematic export of RoboHaMeR's recovered MuJoCo pose: seven arm-joint and 16 modeled hand-joint coordinates per side. All 46 values are MuJoCo hinge-joint positions in radians. Per hand, the 16 coordinates comprise thumb metacarpal, proximal, distal, and tip joints, followed by proximal, distal, and tip joints for each of four fingers. Six hand joints per side are independently actuated. The five tip fields per hand are fixed at zero in the MuJoCo model (zero-based 46D indices 10, 13, 16, 19, 22, 33, 36, 39, 42, 45). In episode 0 (97 frames), \path{observation.state[t]} and \path{action[t]} are identical at every frame: both store the reconstructed pose $q_t$; the \path{action} column represents it as an absolute joint-position command.

To support reproducibility, we distinguish between the released 46D kinematic dataset and the 13D control representation used for policy training. The evaluated 13D trajectories were generated directly from the reconstructed 61D MuJoCo poses, using next-frame absolute joint targets. Both representations derive from the same 340 generated episodes. The 13D channels contain seven right-arm joint positions in radians and six right-hand actuator targets normalized to their joint ranges. The left side is absent from the policy state and action vectors. The 13D export uses native generated RGB video; it is not a truncation of the 46D Parquet arrays. Let $x_t$ denote its 13D vector at frame $t$. The export stores $\mathrm{state}[t]=x_t$ and $\mathrm{action}[t]=x_{t+1}$ at 15\,Hz, omitting the final row of each episode. GR00T predicts 16 future absolute joint targets from these video-reconstructed pseudo-labels. Across 340 episodes, this produces 32,000 rows; the evaluated policy used the 300 accepted episodes. Code also defines a compact bilateral 26D control representation (seven arm and six independently actuated hand coordinates per side). No 26D policy was evaluated in the reported robot trials.

\section{Preliminary Results}
\label{sec:results}

\begin{figure}[t]
    \centering
    \includegraphics[width=1.0 \linewidth]{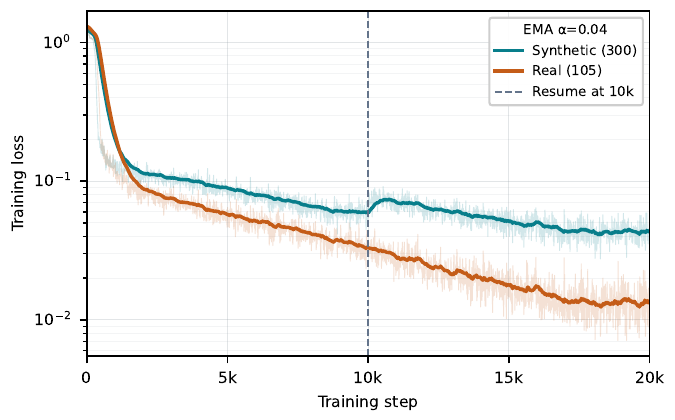}
    \caption{Training loss of separate GR00T~N1.6 policies on
    300 synthetic episodes across eight objects or 105 real Oreo episodes
    (20k steps each). Thick curves show smoothed losses. The dashed line
    marks the resumption of the synthetic run at step 10k. Training loss
    measures the optimization objective, while robot performance is
    evaluated in the pick-and-place trials.}
    \Description{Log-scale training loss over 20,000 steps for separate
    GR00T N1.6 policies trained on synthetic and real demonstrations.}
  \label{fig:results}
\end{figure}

As a preliminary assessment of whether the generated action labels provide useful supervision, we fine-tuned a pretrained GR00T~N1.6 policy for 20,000 steps exclusively on the 300 quality-screened synthetic episodes, without including real demonstrations in the fine-tuning dataset. We then deployed this specific fine-tuned right-side 13D checkpoint for inference on a real-world pick-and-place task with an Oreo package, where it achieved 16/20 (80\%) success across 20 trials. Each trial began from a canonical initial pose (arms bent at the elbows, wrists above the table) with a small random perturbation to the joint configuration. The object was randomly placed within a 15\,cm radius of the robot's operating hand. A trial was scored as successful if the robot grasped the Oreo package and successfully released it into the target plate within a 30-second timeout; no retries or operator interventions were permitted once the trial started. For comparison, we fine-tuned a separate GR00T~N1.6 policy for 20,000 steps on 105 real teleoperated Oreo episodes, using the same right-side 13D layout and next-frame action targets. Under the same 20-trial robot protocol, it achieved 18/20 (90\%) successes, compared with 16/20 (80\%) for the synthetic-data policy. Figure~\ref{fig:results} shows the training losses of both policies.

\emph{Compute cost.} All training steps are one-time machine costs: RoboHaMeR extractor 5h (40GB GPU), Cosmos-3 transfer LoRA 5h (60GB GPU), Cosmos-3 video LoRA $\sim$1 day (60GB GPU), GR00T~N1.6 fine-tuning 3h (45GB GPU). Generating additional synthetic clips reuses the trained models without further teleoperated demonstrations.

\section{Discussion and Limitations}
\label{sec:discussion}

The pipeline has two potentially interacting sources of error: the Cosmos-3 video model may generate implausible motion outside its fine-tuning distribution, and generator artifacts can in turn affect RoboHaMeR's ability to accurately recover the underlying motion of the generated clip. Under the same Oreo protocol, the real-only policy succeeded in 18/20 trials versus 16/20 for the synthetic-only policy. The two-success difference is descriptive: each policy was tested in only 20 trials, and the training sets differ in size and task composition. Scene diversity is limited to eight object categories and one robot/camera setup. Further trials, a zero-shot control, and broader object coverage are needed.

No controlled ablations isolate Scene Recomposition or compare
alternative synthesis strategies. The evaluated policy used the
300 accepted episodes; without a matched unfiltered policy, its
success does not isolate the effect of quality filtering.

The absence of the full generation and action-extraction code and
their execution configurations limits independent reproduction of the pipeline.

Technical utility was assessed in one autonomous tabletop task.
Assessing effects on collection time, operator fatigue, and cognitive
load requires a comparative study accounting for seed collection,
setup, and quality review.

\section{Conclusion}
\label{sec:conclusion}

We presented DataForgeVLA, a pipeline for generating fully synthetic video--action demonstrations. On one Oreo task, GR00T~N1.6 trained on 300 accepted synthetic episodes achieved 16/20 (80\%) successes, versus 18/20 (90\%) for a real-only policy under the same 20-trial protocol. Broader utility requires more objects, trials, and baselines.

\section*{Ethics and Privacy Statement}

Teleoperated demonstrations seed synthetic videos of inanimate-object manipulation. No human participants were evaluated. The pipeline could be misused to generate deceptive robot video; use requires a calibrated robot rig and access to its pose data.

% Start balancing in the first column of the final page.
\balance
\bibliographystyle{ACM-Reference-Format}
\bibliography{refs}

\end{document}